\documentclass[runningheads]{llncs}

\usepackage[mobile]{eccv/eccv}

\usepackage{eccv/eccvabbrv}

\usepackage{graphicx}
\usepackage{booktabs}

\usepackage[accsupp]{axessibility}  %
\usepackage[pagebackref,breaklinks,colorlinks,citecolor=eccvblue]{hyperref}
\usepackage{orcidlink}

\newcommand{\methodName}{HeadGaS}
\newcommand{\methodNameSpace}{HeadGaS }

\usepackage{multirow}
\usepackage{amsmath,bm}

\begin{document}

\title{\methodName: Real-Time Animatable Head Avatars via 3D Gaussian Splatting} 

\titlerunning{\methodName}

\author{Helisa Dhamo
\and Yinyu Nie
\and Arthur Moreau
\and Jifei Song
\and Richard Shaw
\and \\ Yiren Zhou
\and Eduardo P\'erez-Pellitero
}

\authorrunning{H.~Dhamo et al.}

\institute{Huawei Noah's Ark Lab\\
\email{e.perez.pellitero@huawei.com}}

\maketitle
\begin{figure}[h]
\centering
\includegraphics[width=0.53\linewidth]{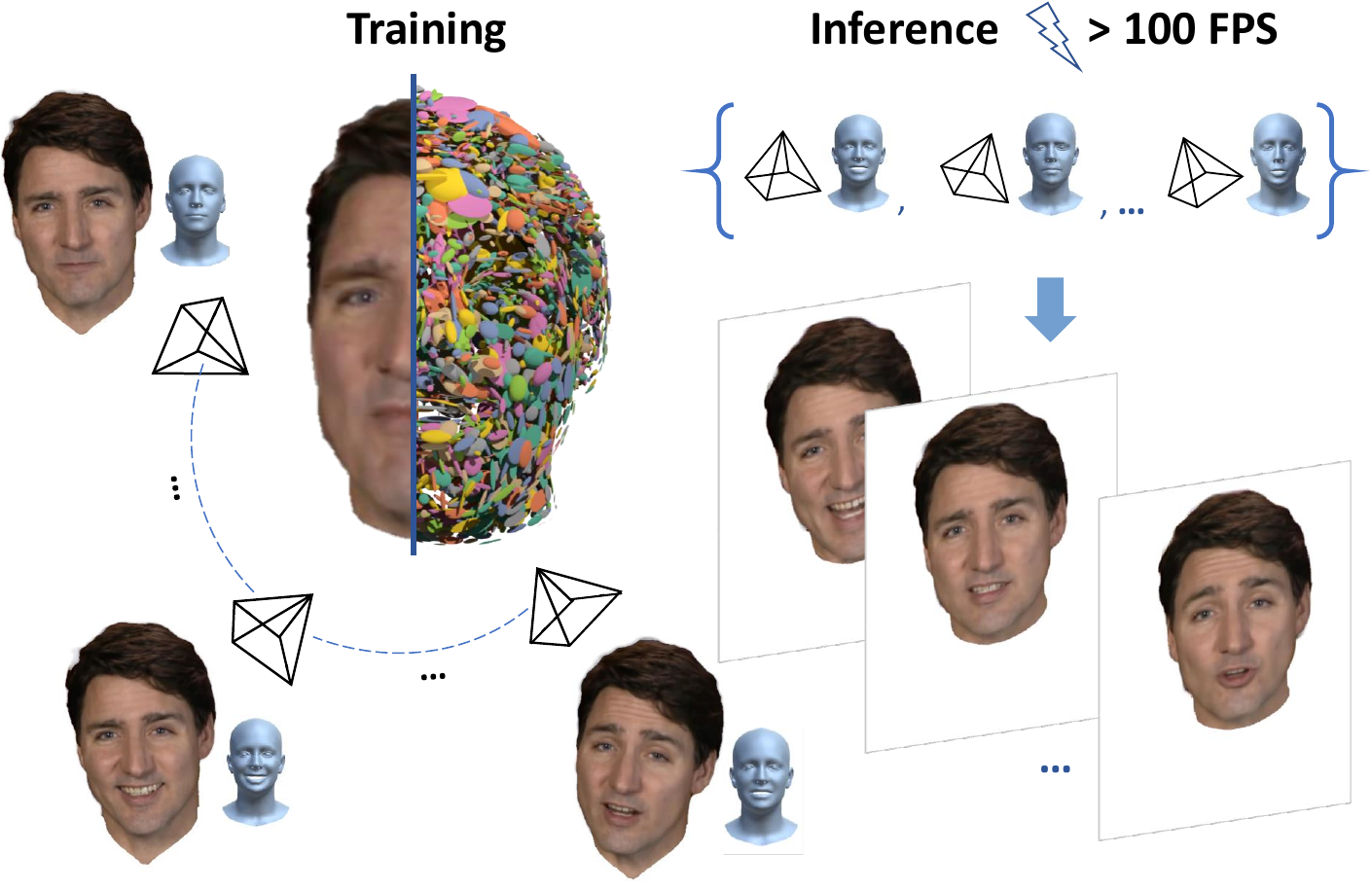}\includegraphics[width=0.43\linewidth]{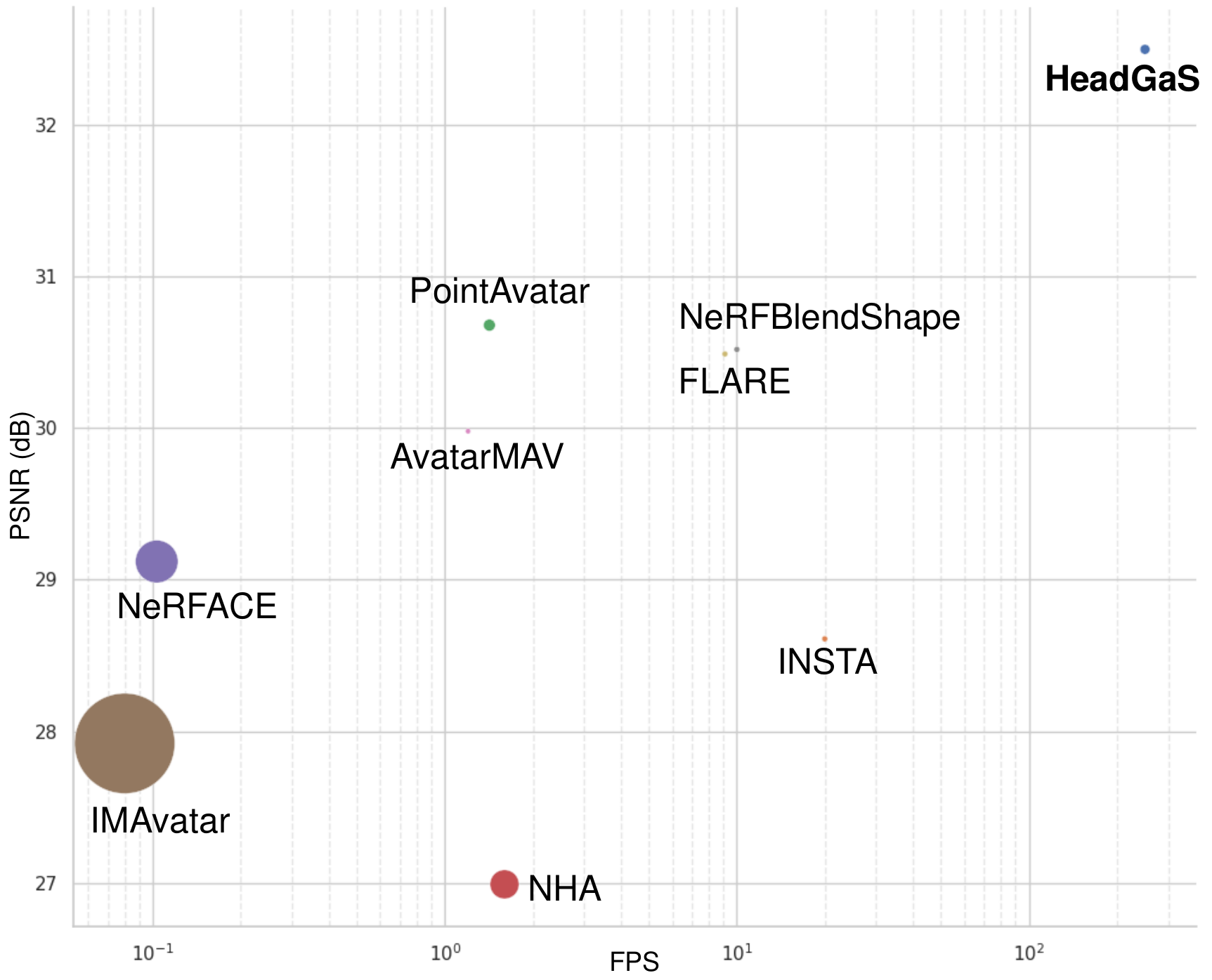}
    \caption{\textbf{Overview of \methodName.} We reconstruct a 3D head based on an expression-aware 3D Gaussian cloud representation, which results in real-time rendering and high image quality. \textbf{Left:} The model is trained with a monocular video of a moving head. At inference, we query the model with a novel sequence of poses and expression parameters to render a real-time video. \textbf{Right:}  Rendering speed (fps in logarithmic scale) vs PSNR plot comparing different methods. The circle radius indicates training time.}
\label{fig:teaser}
\end{figure}
\vspace{-1cm}

\begin{abstract}
    3D head animation has seen major quality and runtime improvements over the last few years, particularly empowered by the advances in differentiable rendering and neural radiance fields. Real-time rendering is a highly desirable goal for real-world applications. We propose HeadGaS, a model that uses 3D Gaussian Splats (3DGS) for 3D head reconstruction and animation. In this paper we introduce a hybrid model that extends the explicit 3DGS representation with a base of learnable latent features, which can be linearly blended with low-dimensional parameters from parametric head models to obtain expression-dependent color and opacity values. We demonstrate that HeadGaS delivers state-of-the-art results in real-time inference frame rates, surpassing baselines by up to $2$ dB, while accelerating rendering speed by over $\times 10$.
\end{abstract}

\section{Introduction}
\label{sec:intro}
Reconstructing photorealistic 3D heads which are in turn controllable and naturally expressive is essential for building digital avatars that look and behave like real humans. This has a wide range of applications including AR/VR, teleconferencing, and gaming. Designing head models that accomplish high fidelity in their appearance, are easy to capture and enable expressive control has been an active research field in recent years, specially due to the fast development of neural and differentiable rendering approaches.

Animatable 3D head reconstruction consists in driving a captured head avatar, based on a target sequence of facial expressions and head poses. %
In the last decades, various parametric 3D morphable models (3DMM) have emerged \cite{10.1145/311535.311556,FLAME:SiggraphAsia2017,facewarehouse}, which can be fitted to sequences of a moving head and later on enable  pose and expression control. Though these models make it possible to drive a captured avatar via a set of low-dimensional parameters, generally their generated images lack realism. Other works utilize the fitting of low-dimensional parameters from such 3DMM models for initial estimates and build on other mechanisms to obtain more realistic imagery with animation capabilities~\cite{grassal2021neural,Gafni_2021_CVPR}.  

In particular, with the recent success of differentiable rendering, various 3D-aware animatable head models emerged that can reconstruct and render 3D heads, while providing the functionality to drive them based on expression parameters from 3DMM models. These representations can be explicit (mesh, point clouds)~\cite{grassal2021neural,Zheng2023pointavatar} or implicit (neural)~\cite{Gao2022nerfblendshape}. Thereby, the explicit models impose stronger constraints on the head surface, which allows for better expression and pose generalization, while making it more difficult to preserve photo realism, as they inherit the limitations and artifacts of the underlying representation (mesh, point cloud) %
as observed in Gao \etal~\cite{Gao2022nerfblendshape} and also seen in our experiments (Fig.~\ref{fig:result}).  %
With the recent success of neural radiance fields (NeRFs)~\cite{mildenhall2020nerf}, typically implicit models are based on a NeRF representation~\cite{Gafni_2021_CVPR}. Some of these models~\cite{INSTA:CVPR2023,Gao2022nerfblendshape} prioritize time constraints and therefore rely on very fast volumetric NeRF variants (\eg InstantNGP~\cite{mueller2022instant}) to enable fast training and rendering.

Despite impressive efforts to improve NeRFs to be more accurate~\cite{barron2023zipnerf} and fast~\cite{mueller2022instant}, there is a trade-off between these two aspects that is hard to satisfy simultaneously~\cite{barron2023zipnerf}. Moreover, even fast and efficient NeRF models like InstantNGP typically enable interactive inference frame rates at best (10-15 fps)~\cite{kerbl3Dgaussians}. Very recently, 3D Gaussian Splatting (3DGS)~\cite{kerbl3Dgaussians} emerged as a competitive alternative to NeRF, which leads to reasonable photo-realism while bringing the rendering speed to real-time rates. This is thanks to its representation as a set of 3D Gaussian primitives, with a more efficient space coverage compared to point clouds, combined with efficient tile-based rasterization. However, in the light of 3D head animation, in its original form, 3DGS does not constitute an intuitive surface or point set that can be directly deformed based on 3DMM deformation, unlike other well-known representations, \eg surface or pointcloud based. %

\begin{figure}[t]
    \centering
    \includegraphics[width=0.94\linewidth]{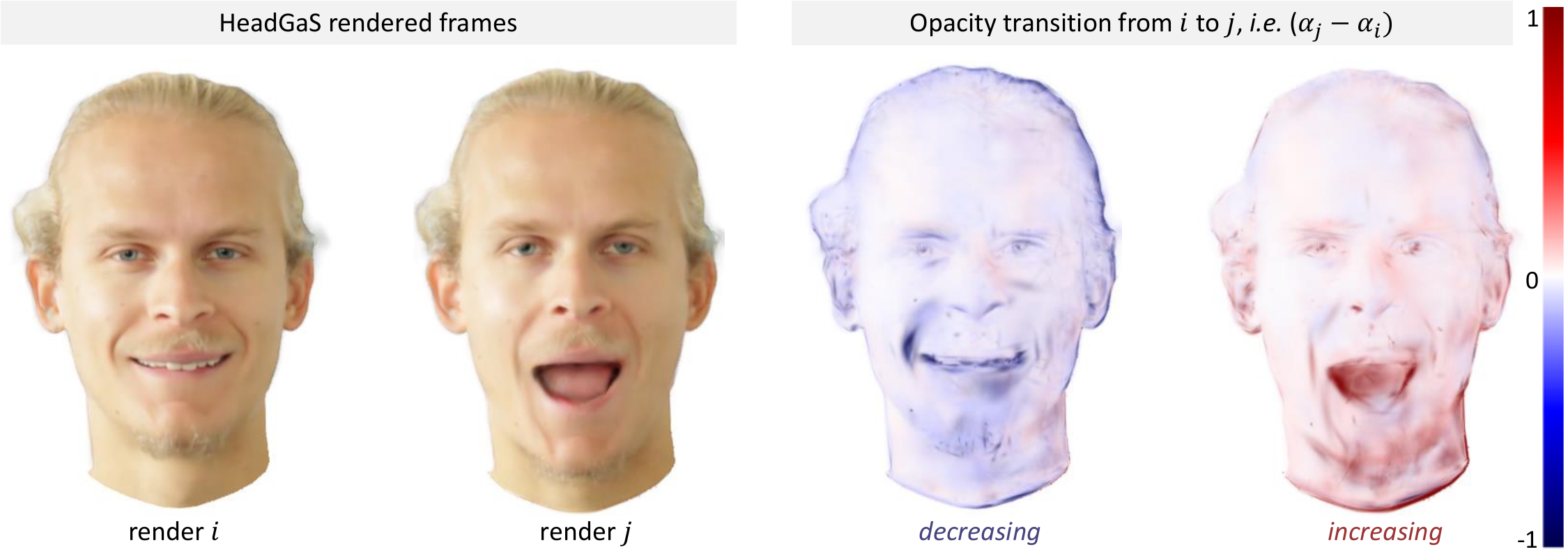}
    \caption{\textbf{Motion modelling via opacity change.} \textbf{Left}: Two example frames $i$ and $j$ rendered by \methodName. \textbf{Right}: Rendering of opacity difference $\alpha_j - \alpha_i$ (\textcolor{blue}{blue}: Gaussians with an opacity decrease; \textcolor{red}{red}: Gaussians with an opacity increase; \textit{colors close to white}: minor change, static regions). We observe a strong opacity increase in dynamic areas, \eg lower chin Gaussians turn opaque as the jaw fully opens.}
    \label{fig:idea_overview}
\end{figure}

To circumvent this limitation, we propose \methodName, a model that enhances 3D Gaussians with head animation capabilities (see Figure~\ref{fig:teaser}). At test time, our model receives a sequence of head poses and expression parameters, and generates a photo-realistic video of the reconstructed avatar. The core idea behind \methodNameSpace is to represent motion by allowing Gaussians to alter their opacity and color over time. As a consequence, \methodNameSpace will result in duplicates of the dynamic face areas, \ie achieving dynamics via over-representation. Thus, there will be multiple Gaussians representing the same face region, and these duplicates will become active (apparent) one at a time, to support the face geometry at a certain state of expression. Figure \ref{fig:idea_overview} illustrates the opacity change as a result on an expression transition. Note how Gaussians representing lower lip and chin areas in frame $i$ turn transparent to allow for mouth opening in frame $j$, while another set of chin Gaussians emerge at a lower location to accommodate the new jaw position in frame $j$. To allow for such varying appearance, guided by an expression vector, we introduce a basis of latent features inside each Gaussian. This learned basis is multiplied with an input expression vector, and its sum is fed to a multi-layer perceptron (MLP) to yield the final color and opacity. This idea is inspired by traditional blendshape 3DMMs~\cite{facewarehouse}, and it can be interpreted as a latent-feature shape basis that is blended in the feature domain rather than directly in PCA space~\cite{10.1145/311535.311556} or meshes~\cite{facewarehouse}. %
Our model is simple and effective, and it can work with various 3DMM representations, as it does not explicitly model deformations with respect to a particular mesh topology. Practically, in our experiments we show that \methodNameSpace can be controlled with expression parameters from two different 3DMMs, namely FLAME~\cite{FLAME:SiggraphAsia2017} and FaceWarehouse~\cite{facewarehouse}. The rendering is done in real-time framerates, at over 100~fps (about $250$ fps for $512^2$ resolution). %
We show experimentally that our visibility-varying Gaussians outperform the evident alternative of moving the Gaussians, which we attribute to the fact that adding 3D motion makes the optimization even more complex. %

We evaluate our model on publicly available monocular video datasets, commonly used in related works \cite{Gao2022nerfblendshape,INSTA:CVPR2023,Zheng2023pointavatar}.
Thereby, we demonstrate that the proposed model yields superior results, while increasing the rendering speed by at least a $\times 10$ factor compared to interactive NeRF-based baselines~\cite{Gao2022nerfblendshape,INSTA:CVPR2023}. We show the applications of \methodNameSpace in novel same-person expression transfer, cross-subject expression transfer, as well as novel view synthesis. 

To summarize, our contributions include: 
\begin{enumerate}
    \item We formulate a novel framework that can render photo-realistic 3D-aware animatable heads in real-time, adapting an efficient set of 3D Gaussian primitives. This framework handles face dynamics by allowing opacity and color to change over time, \ie leveraging over-representation.
    \item We extend 3DGS~\cite{kerbl3Dgaussians} with a per-Gaussian basis of latent features, which can be blended with expression weights to enable expression control. 
    \item We extensively evaluate our proposed method, and compare it against state-of-the-art approaches, obtaining up to 2dB improvement and $\times 10$ speed-ups.
\end{enumerate}

\section{Related Work}
\label{sec:related}

\subsection{Towards Fast and Dynamic Radiance Fields}
NeRF~\cite{mildenhall2020nerf} represent the scene as an implicit neural radiance field, that queries 3D space and predicts density and view-dependent color via a MLP. In the following years, many follow-up works have focused on improving different aspects of it such as anti-aliasing ~\cite{Barron_2021_ICCV,barron2022mipnerf360,barron2023zipnerf}, regularization for sparse views~\cite{Niemeyer2021Regnerf,kangle2021dsnerf,sparf2023} and speed ~\cite{SunSC22,mueller2022instant,Chen2022ECCV}; or enhancing results post-rendering~\cite{catleychandar2024roguenerf}.  
DVGO~\cite{SunSC22} replace the MLP of NeRF with a density and learned feature voxel grid to considerably speed up convergence. 
TensoRF~\cite{Chen2022ECCV} factorize the 4D feature voxel grid of a scene into a set of low-rank 2D and 3D tensors which improves efficiency. InstantNGP~\cite{mueller2022instant} employ a hash grid and an occupancy grid to accelerate computation, followed by a small MLP that infers density and color.
NeRFs have also been used to represent dynamic scenes including human bodies~\cite{weng_humannerf_2022_cvpr,chen2021animatable,peng2021animatable}, human heads~\cite{Gao2022nerfblendshape,INSTA:CVPR2023}, and generic time-varying scenes~\cite{pumarola2020d,park2021hypernerf,li2020neural,du2021nerflow,park2021nerfies,tretschk2021nonrigid}. Typically these models rely on a canonical space, where all observations are mapped for time consistent reconstruction. Methods aiming at fast rendering speeds ~\cite{INSTA:CVPR2023,Gao2022nerfblendshape} build on an InstantNGP hash grid and achieve interactive frame rates (10-15 fps).

3DGS~\cite{kerbl3Dgaussians} represent a scene as a set of explicit 3D Gaussians with the motivation to minimize computation in empty spaces. Their efficient representation combined with tile-based rasterization algorithm allows for accelerated training and real-time rendering (over 100 fps). A line of works extends the 3D Gaussian representation to model dynamic scenes \cite{luiten2023dynamic,wu20234d,yang2023deformable,Shaw2023SWAGSSW}. Luiten \etal~\cite{luiten2023dynamic} propose simultaneous dynamic scene reconstruction and 6-DoF tracking by allowing the Gaussians to change position and rotation over time while enforcing the same color, size and opacity. Yang \etal~\cite{yang2023deformable} learn a MLP based deformation that maps 3D Gaussians to a canonical space. 4D-GS~\cite{wu20234d} propose an efficient deformation field by querying features in shared multi-resolution voxel planes.  

Very recent concurrent works model human head~\cite{xu2023gaussianheadavatar,xiang2024flashavatar,chen2023monogaussianavatar} and body~\cite{moreau2024human,hugs} avatars with 3DGS, by deforming a canonical head via a 3DMM-conditioned MLP \cite{xu2023gaussianheadavatar,xiang2024flashavatar}, relying on a tri-plane \cite{wang2024gaussianhead}, or binding 3D Gaussians in a FLAME mesh \cite{qian2023gaussianavatars}. Our method is quite different from these works, in that we utilize a per-Gaussian feature basis and opacity induced dynamics.

\subsection{Head Reconstruction and Animation}

Head reconstruction from a set of image observations has been a very active field in the recent years, including models that generalize across subjects~\cite{Mihajlovic:ECCV2022,hong2021headnerf,DBLP:conf/siggraph/WangCZBG22}, or rely on multi-view head captures~\cite{kirschstein2023nersemble,Jang_2023_ICCV,lombardi2018,lombardi2021}, which can have a static~\cite{DBLP:conf/siggraph/WangCZBG22,Jang_2023_ICCV} or dynamic form~\cite{hong2021headnerf,kirschstein2023nersemble}. %
Most related to our proposed method, are works that learn a dynamic, animatable 3D head model from a monocular video sequence that observes the head in various poses and facial expressions, and is capable to generate a novel expression or pose at test time. A line of works rely on explicit scene representations, such as meshes or point clouds~\cite{kim2018deep,Garrido2014AutomaticFR,grassal2021neural,Zheng2023pointavatar}. Neural Head Avatars~\cite{grassal2021neural}  models the geometry and texture explicitly via a hybrid representation consisting of a coarse morphable model, followed by a neural based refinement that predict voxel offsets. %
PointAvatar~\cite{Zheng2023pointavatar} propose a deformable point-based representation, where all points have the same radius. Our 3D Gaussian set shares some similarity with a point cloud, yet it is more flexible. Each Gaussian can have its own radius, orientation and different axis lengths. Recently, FLARE~\cite{bharadwaj2023flare} was proposed, a model that updates a traditional graphics pipeline with a few neural components. FLARE can optimize a 3D mesh via differentiable rendering, enabling avatars that are animatable and relightable. %

Another line of works extend implicit neural radiance representations.
NerFACE~\cite{Gafni_2021_CVPR} use a dynamic NeRF to combine scene information with a morphable head model to enable pose and expression control. IMAvatar~\cite{zheng2022imavatar} utilizes neural implicit surfaces~\cite{Park_2019_CVPR} and learns an implicit deformation field from canonical space to observation based on expression parameters and pose.
With the goal of fast training, and interactive rendering, more recent works extend InstantNGP~\cite{mueller2022instant} with head aware models. INSTA~\cite{INSTA:CVPR2023} use a tracked FLAME mesh as a geometrical prior to deform the points into canonical space, followed by InstantNGP~\cite{mueller2022instant}. NeRFBlendShape~\cite{Gao2022nerfblendshape} follow a different approach, that in contrast to most previous works does not rely on deformation. Instead, inspired by classic blendshape models for heads~\cite{facewarehouse}, they utilize a base of multi-level hash grid fields~\cite{mueller2022instant}, where the model can be driven via a linear blending of such hash grid base with the expression vector. Similarly, AvatarMAV~\cite{xu2023avatarmav} use expression weights to blend a set of motion voxel grids.
Our proposed approach resembles these blending-based ideas, but instead we use the expression vector to blend latent per-Gaussian features to predict expression-specific color and opacity.

Different from all works discussed in this section, we adopt a set of 3D Gaussians as neural radiance representation, to take advantage of the fast rendering benefits, combined with competitive photorealism.

\begin{figure}[t]
    \begin{center}
        \includegraphics[width=\linewidth]{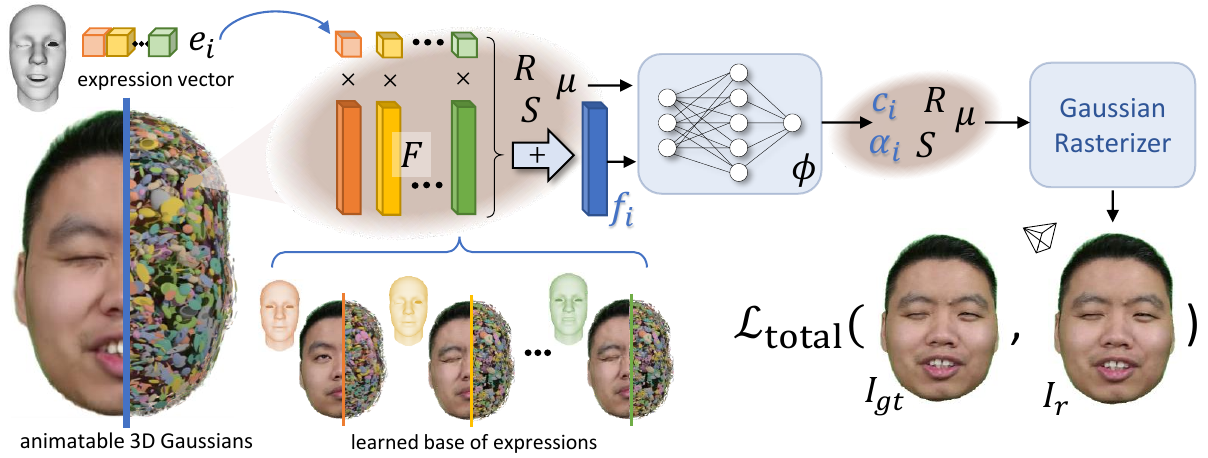} 
    \end{center}
    \caption{\textbf{\methodNameSpace pipeline.} We represent 3D space as a set of feature-enhanced 3D Gaussians. Every Gaussian contains a feature basis $\bm{F}$ that can be blended via the expression vector to obtain a frame specific feature $\bm{f}_i$. The frame specific feature is fed to an MLP $\phi(\cdot)$ alongside position $\bm{\mu}$ to obtain expression-dependent color $\bm{c}_i$ and opacity $\bm{\alpha}_i$. Finally, $\bm{c}_i$ and $\bm{\alpha}_i$ are fed to the rasterizer alongside other Gaussian parameters like rotation $R$, scale $S$ and position $\bm{\mu}$ to render the image.}
    \label{fig:method}
\end{figure}

\section{Method}
\label{sec:method}

Given a monocular video of a moving head, our goal is to learn a 3D head model and render novel images of this avatar based on a facial expression vector and camera pose. As a pre-processing step, similar to other works~\cite{Gao2022nerfblendshape, INSTA:CVPR2023,grassal2021neural,Zheng2023pointavatar}, we require head poses and a vector of expression weights associated with each frame, and adopt a head tracking pipeline to achieve this. Note that \methodNameSpace does not explicitly build on a certain parametric model, and it can therefore work with different head models. We have performed experiments with two models~\cite{FLAME:SiggraphAsia2017,facewarehouse} based on the head tracking frameworks of prior work~\cite{Gao2022nerfblendshape,INSTA:CVPR2023,zheng2022imavatar}. The resulting rigid head poses are converted to camera poses, to map all observations to the canonical head pose. In addition, we perform video matting~\cite{Lin_2022_WACV} followed by face parsing~\cite{faceparsing} to discard background areas and clothing to focus on the head region only. In the following sections, we will first provide a background on the original 3DGS (Sec.~\ref{sec:gaussian}) %
and further describe the proposed strategy for animatable 3DGS (Sec.~\ref{sec:bs}), the rendering (Sec.~\ref{sec:rendering}) and its optimization (Sec.~\ref{sec:optim}). 

\subsection{Original 3DGS Representation}
\label{sec:gaussian}

Given a set of images of a static scene and the corresponding camera poses, 3DGS~\cite{kerbl3Dgaussians} learn a 3D scene as a set of 3D Gaussians, and can render a novel image from a given viewpoint. For initialization, 3DGS utilizes a sparse point cloud, typically originating from the COLMAP \cite{schoenberger2016sfm} framework by which they also obtain the camera poses. Thereby, a 3D Gaussian is represented as a tuple of 3D covariance matrix $\bm{\Sigma} \in \mathbb{R}^{3\times3}$, Gaussian center $\bm{\mu} \in \mathbb{R}^3$,
color $\bm{c} \in \mathbb{R}^{3(k+1)^2}$ and opacity $\alpha \in \mathbb{R}$, \ie $\mathcal{G} = (\bm{\Sigma}, \bm{\mu}, \bm{c}, \alpha)$, where $k$ is the degree of the spherical harmonics. The Gaussians are defined in world space, centered at the mean point
\begin{equation}
    G(x) = e^{- \frac{1}{2}(x)^T\bm{\Sigma}^{-1}(x)}.
\end{equation}
To make optimization stable, \ie \ guarantee that $\bm{\Sigma}$ is positive semi-definite, the covariance matrix is further decomposed into rotation $R$ and scaling $S$:
\begin{equation}
    \bm{\Sigma} = RSS^{T}R^{T}.
\end{equation}
Color is given as spherical harmonics (SH) of degree $k$ and is thus view-dependent.

The Gaussian parameters are optimized via a differentiable rasterizer, that projects the current 3D Gaussians to the image space and compares against the ground truth images. 
This rasterizer relies on an efficient algorithm for sorting the Gaussians and tiling the image space, which leads to very fast training and rendering. Alongside the optimizations, 3DGS employs an adaptive mechanism for pruning and densification, to make sure that the set of gaussians represents the space effectively. For more details, we refer the reader to the 3DGS paper~\cite{kerbl3Dgaussians}.

\subsection{Feature Blending Formulation}
\label{sec:bs}

Here we describe how we extend the 3DGS representation with animation capabilities. The vanilla 3DGS model does not inherently allow for this, as it learns a static set of parameters, which is the same for all frames. Inspired by 3DMMs, the goal of our model is to explore a blending mechanism for the 3D Gaussian components, using the pre-computed expression parameters as blending weights. Namely, we want each Gaussian to change color and opacity based on the current expression $i$. This leads to 3D Gaussians with dynamic appearance which occasionally appear and vanish depending on the current expression, and additionally allow color changes for non-rigid appearance effects such as wrinkles. For instance, referring to Figure \ref{fig:idea_overview}, Gaussians of closed lips visible in frame $i$ will turn transparent in frame $j$, as the jaw opens, while another set of Gaussians at a \emph{different} location will become visible to render open lips. The model will thus learn multiple Gaussians corresponding to the same region in the face, such that these can become opaque as needed. %

With the goal of enabling such dynamic appearance, we extend \textit{every} 3D Gaussian with a basis of latent features $\bm{F} \in \mathbb{R}^{B \times f_{dim}}$, (see Figure~\ref{fig:method}). Our animatable 3D Gaussian representation then becomes $\mathcal{G}_a = (\bm{\Sigma}, \bm{\mu}, \bm{F})$. 
The latent base is optimized together with the other parameters of the 3D Gaussian. At each iteration, we leverage the respective expression weights $\bm{e}_i \in \mathbb{R}^{B}$ of the current frame $i$, to blend the feature basis $\bm{F}$ into a 1D vector $\bm{f}_i \in \mathbb{R}^{f_{dim}}$

\begin{equation}
 \bm{f}_i = \bm{F}^{T} \bm{e}_i + \bm{f}_0   
\end{equation}
where $\bm{f}_0$ is a bias term.
We index with $i$ all variables that are specific to a particular frame $i$.
This frame specific feature $\bm{f}_i$ is then fed into a small MLP $\phi(\cdot)$, to compute the color $\bm{c}_i$ as well as the opacity $\alpha_i$
\begin{equation}
    \bm{c}_i, \alpha_{i} = \phi(\bm{f}_i, \psi(\bm{\mu}))
\end{equation}
where $\psi$ denotes sinusoidal positional encoding, the learned color is a 1D vector $\bm{c}_i \in \mathbb{R}^{3(k+1)^2}$, and the learned opacity is a scalar $\alpha_i \in \mathbb{R}$. As most of the dynamic effects are already captured by the per-Gaussian feature bases $\bm{F}$, we are able to use a very small MLP that does not compromise the rendering speed. Our MLP is composed of only two linear layers, each followed by LeakyReLU activation~\cite{xu2015empirical}, where the hidden layer has $64$ channels. The last layer consists of two branches, \ie for color and opacity prediction. We use a sigmoid activation function at the end of the opacity branch to constrain it to be in its appropriate range $[0,1]$. As color and opacity are learned via the MLP, we omit them from the explicit optimizable  Gaussian parameters. 

An alternative to blending in the latent space, would be to directly define a basis of explicit Gaussian parameters and similarly blend them based on the expression weights. %
However, as these values have an explicit meaning (\ie color, position), a multiplication with expression weights that are not even learnable, makes this formulation limiting and prone to artifacts, as we also show in our ablation (Sec.~\ref{sec:ablations}, \textit{Ours w/o MLP}).
Interestingly, even though changing the centers and rotations of the Gaussian splats (instead of modifying opacity and colour) would be an intuitive mechanism when it comes to 3D Gaussian animation, our proposed approach results in much better performance (see Sec.~\ref{sec:ablations}, \textit{Ours w/ $\Delta(\mu,R)$}). We additionally show that the proposed feature blending strategy is superior to the straight-forward approach of using the expression vector as a condition to the MLP (\textit{Ours w/o blending}). %

\subsection{Rendering}
\label{sec:rendering}

To render frame $i$, we employ the respective expression $\bm{e}_i$ to populate each Gaussian with expression-dependent color and opacity. Then, the Gaussians are rendered using the camera view $\bm{W}_i$, and similarly to Kerbl \etal~\cite{kerbl3Dgaussians} we perform the splatting technique on the primitives~\cite{Yifan:DSS:2019}. Given a viewing transform $\bm{W}$ as well as the Jacobian of the affine approximation of the projective transformation $\bm{J}$, the covariance matrix $\bm{\Sigma}'$ in camera coordinates can be obtained from
\begin{equation}
    \bm{\Sigma}' = \bm{J}\bm{W} \bm{\Sigma} \bm{W}^T \bm{J}^T.
\end{equation}
The Gaussian splats are then rendered via a tile-based differentiable rasterizer~\cite{kerbl3Dgaussians} that pre-sorts all primitives of an image at once.

\subsection{Optimization}
\label{sec:optim}

We initialize the 3D Gaussians centers with 2500 points. Whenever available, these points are a subset of vertices from the tracked 3DMM meshes (\eg FLAME based data released by prior works~\cite{INSTA:CVPR2023,Zheng2023pointavatar}). As there is no mesh available for the data from Gao \etal~\cite{Gao2022nerfblendshape}, we sample random points within the given near and far bounds. Empirically we found that initializing the latent features $\bm{F}$ with zeros led to the most stable solution.
The model is optimized by rendering the learned Gaussians and comparing the resulting image $I_\text{r}$ against the ground truth $I_\text{gt}$. We minimize the following loss objective

\begin{equation}
    \mathcal{L}_\text{total} = \lambda_{1} \mathcal{L}_{1}(I_\text{r}, I_\text{gt}) + \lambda_{s} \mathcal{L}_\text{SSIM} (I_\text{r}, I_\text{gt}) + \lambda_{p} \mathcal{L}_{p} (I_\text{r}, I_\text{gt})
\end{equation}
where the $\lambda$s are weighting factors and $\mathcal{L}_{p}$ denotes the perceptual loss~\cite{Johnson2016Perceptual}. We optimize using Stochastic Gradient Descent~\cite{ruder2016overview} with a standard exponential decay scheduling for the Gaussian position centers $\bm{\mu}$ as well as the MLP.

\subsubsection{Adaptive densification and pruning}
Following 3DGS~\cite{kerbl3Dgaussians}, we combine our optimization with periodic steps of adaptive densification and pruning. First, this mechanism prunes Gaussians that are almost transparent, \ie $\alpha < \tau_{\alpha}$ smaller than a threshold. Second, the densification targets areas that need to be populated with more Gaussians, represented with Gaussians that are too large, or regions that are too sparse and lack detail. Based on the observation that in both cases the position gradients have high values~\cite{kerbl3Dgaussians}, the Gaussians that should be densified are identified utilizing the average gradient magnitude being above a threshold $\tau_{pos}$. In the case of Gaussians that are too small, the objective is to increase volume and therefore the identified Gaussians are simply cloned, preserving their size. On the other hand, for Gaussians that are too large, the goal is to preserve the overall volume and therefore their scales are decreased by a factor of $1.6$ after cloning, obtained empirically by~\cite{kerbl3Dgaussians}.  %

\subsection{Implementation Details}

The learning rates for the MLP $\phi(\cdot)$, positions $\mu$, latent features $\bm{F}$, scale $S$ and rotation $R$ are namely $1.6{\cdot}10^{\text{-}4}$, $1.6{\cdot}10^{\text{-}4}$, $0.0025$, $0.005$ and $0.001$. We set the latent feature dimensionality to $f_{dim}{=}32$. For the FLAME tracking~\cite{INSTA:CVPR2023} we only use the first 52 expression weights, \ie $B{=}52$. All Gaussians are fed as a single batch into the MLP. The $\mathcal{L}_{p}$ loss is based on a VGG network \cite{Simonyan15} and has a weight of $\lambda_{p}{=}0.1$, while $\lambda_{1}{=}0.8$ and $\lambda_{s}{=}0.2$. We activate the $\mathcal{L}_{p}$ loss after $10k$ iterations such that it does not conflict with photometric loss at the early stage of learning. To save computation, we apply $\mathcal{L}_{p}$ on the image region defined by the head bounding box. The densification starts after 500 iterations and stops with 15k iterations. In our experiments we use an SH degree of $k{=}3$.
We train our models on one Tesla $V100$ GPU for $50k$ iterations taking about 1 hour.

\section{Experiments}
\label{sec:exp}

In this section we describe the evaluation protocol followed by quantitative and qualitative results in three different scenarios, \eg same-subject novel expression and novel view rendering, as well as cross-subject expression driving.

\begin{table*}[t]
    \centering
    \caption{Results on the dataset provided by INSTA~\cite{INSTA:CVPR2023},  NeRFBlendShape~\cite{Gao2022nerfblendshape} (NBS) and PointAvatar~\cite{Zheng2023pointavatar}. We report PSNR, SSIM and LPIPS, together with time measures in seconds of 1 frame rendering (for $512^2$ resolution).}
    \resizebox{0.8\columnwidth}{!}{%
    \begin{tabular}{l|c|cccc|r} \toprule

         Method & dataset & L2 $\downarrow$ & PSNR $\uparrow$ & SSIM $\uparrow$ &  LPIPS $\downarrow$ & Time (s) $\downarrow$ \\
         \midrule
         NHA~\cite{grassal2021neural} & \multirow{9}{*}{INSTA} & 0.0024 & 26.99 & 0.942 & 0.043 & 0.63 \\
          IMAvatar~\cite{zheng2022imavatar} & & 0.0021 & 27.92 & 0.943 & 0.061 & 12.34 \\
          NeRFACE~\cite{Gafni_2021_CVPR} &  & 0.0016 & 29.12 & 0.951 & 0.070 & 9.68 \\
          AvatarMAV~\cite{xu2023avatarmav} & & 0.0012 & 29.98 & 0.948 & 0.079 & 0.85 \\
          FLARE~\cite{bharadwaj2023flare} & & 0.0010 & 30.49 & 0.942 & 0.050  & 0.11 \\
         INSTA~\cite{INSTA:CVPR2023} &  & 0.0017 & 28.61 & 0.944 & 0.047 & 0.05 \\
         PointAvatar~\cite{Zheng2023pointavatar} &  & 0.0009 & 30.68 & 0.952 & 0.058 & 0.1 - 1.5 \\
        NeRFBlendShape~\cite{Gao2022nerfblendshape} &  & 0.0011 & 30.52 & 0.955 & 0.056 & 0.10 \\
         \methodNameSpace (Ours) &  & \textbf{0.0008} & \textbf{32.50} & \textbf{0.971} & \textbf{0.033} & \textbf{0.004} \\
         \midrule
         NeRFBlendShape~\cite{Gao2022nerfblendshape} & \multirow{2}{*}{NBS} & 0.0005 & 34.34  & 0.970  & 0.0311 & 0.10 \\ 
        
         \methodNameSpace (Ours) &  & \textbf{0.0003} & \textbf{36.66} &  \textbf{0.976} & \textbf{0.0261} & \textbf{0.004} \\
         \midrule
         
    PointAvatar~\cite{Zheng2023pointavatar} & \multirow{2}{*}{PointAvatar} & \textbf{0.0027}  & \textbf{26.04} & 0.885 & 0.147 & 0.1 - 1.5 \\
          \methodNameSpace (Ours) &  & 0.0029 & 25.99 & \textbf{0.897} & \textbf{0.108} & \textbf{0.004} \\
          \bottomrule
    \end{tabular}
    }
    \label{tab:result1}
\end{table*}
We evaluate our model on three datasets, made publicly available by recent works such as NeRFBlendShape (we dub this NBS data), INSTA, and PointAvatar.
The NBS data~\cite{Gao2022nerfblendshape} contains a set of monocular videos from 8 subjects, where the last $500$ frames in each subject constitute the test set. The INSTA dataset~\cite{INSTA:CVPR2023} contains 10 subjects, with the last $350$ frames of each sequence being the test set. We additionally evaluate on the 3 subjects made available by PointAvatar~\cite{Zheng2023pointavatar}, where the test sets contain between $880-1800$ frames per subject. For fairness of comparison, when training our method in all three datasets, we use the same splits and utilize the tracked data (head poses and expression weights) provided originally by the authors, such that tracking quality does not affect the comparison. To train our model we use a subset of about $1500-2500$ frames from the training set of each subject.
To assess the quality of the synthesized images we report common metrics such as Peak Signal-to-Noise Ratio (PSNR), Structural Similarity (SSIM), Learned Perceptual Image Patch Similarity (LPIPS)~\cite{zhang2018perceptual} and the Mean Squared Error (L2). All metrics are computed using white background for non-face regions. Further, we report rendering times in seconds. 
We compare against common baselines such as NeRFBlendShape~\cite{Gao2022nerfblendshape}, INSTA~\cite{INSTA:CVPR2023}, PointAvatar~\cite{Zheng2023pointavatar}, NHA~\cite{grassal2021neural}, IMAvatar~\cite{zheng2022imavatar}, AvatarMAV~\cite{xu2023avatarmav}, FLARE~\cite{bharadwaj2023flare} and NeRFACE~\cite{Gafni_2021_CVPR}. The evaluation is carried out using their official code repositories, as well as their official checkpoints whenever available. %

\subsection{Same-Subject Novel Expression Driving}

\begin{figure*}%
    \begin{center}
        \includegraphics[width=0.99\linewidth]{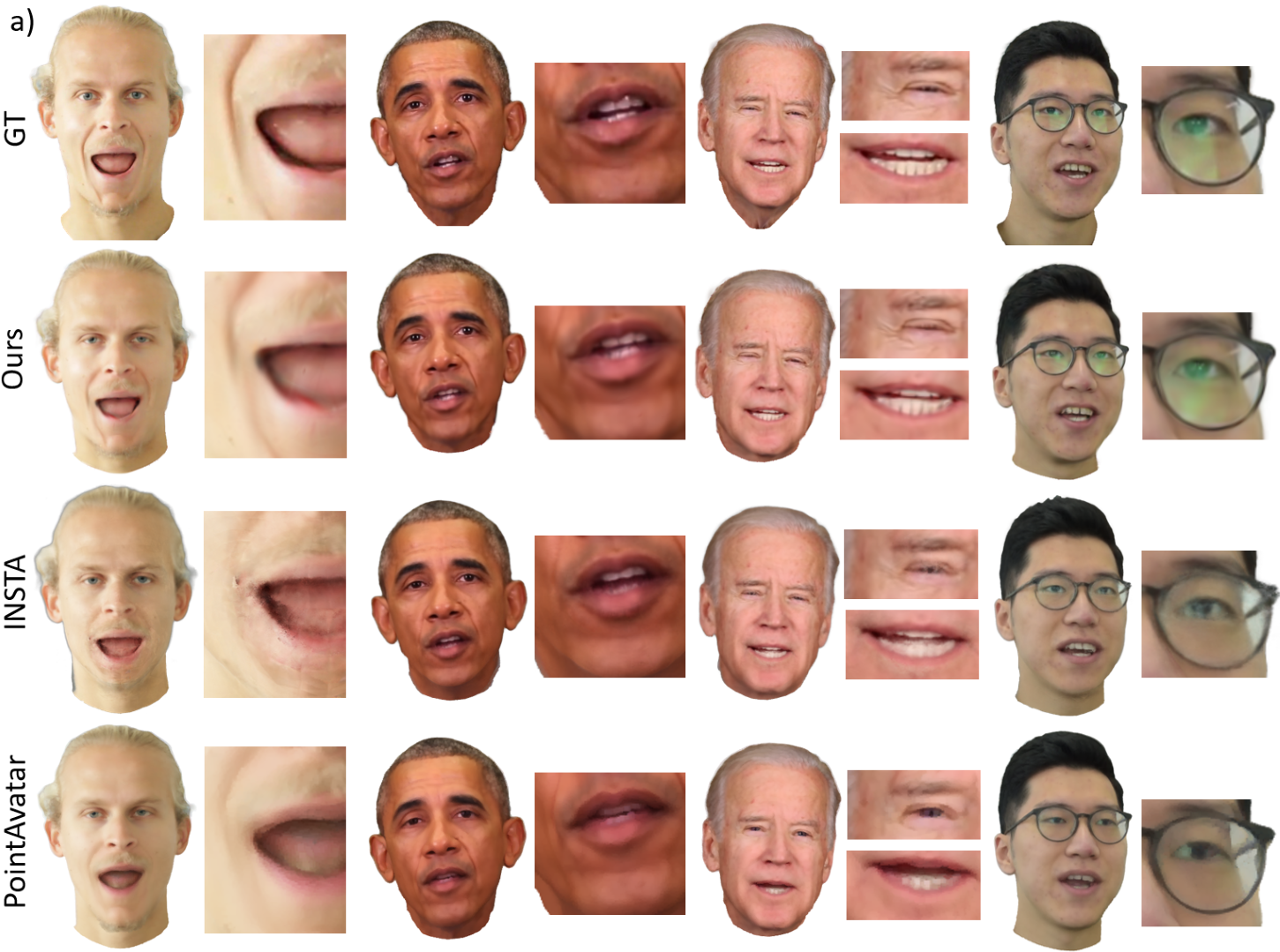} 
        \includegraphics[width=0.99\linewidth]{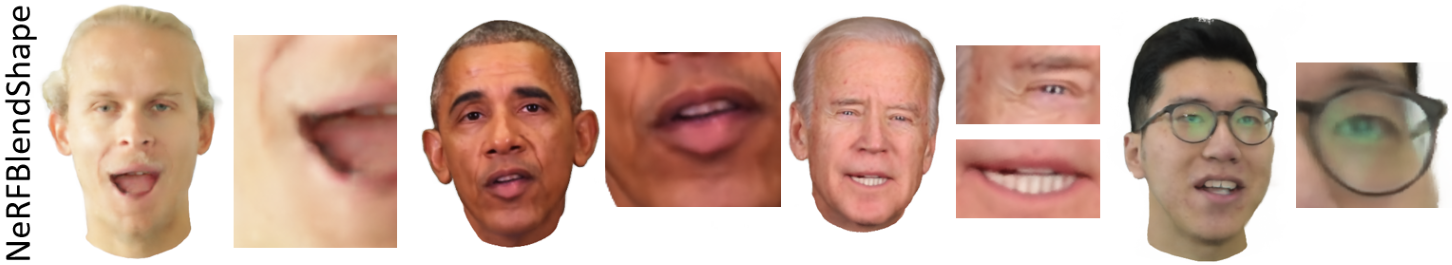} 
        \includegraphics[width=0.98\linewidth]{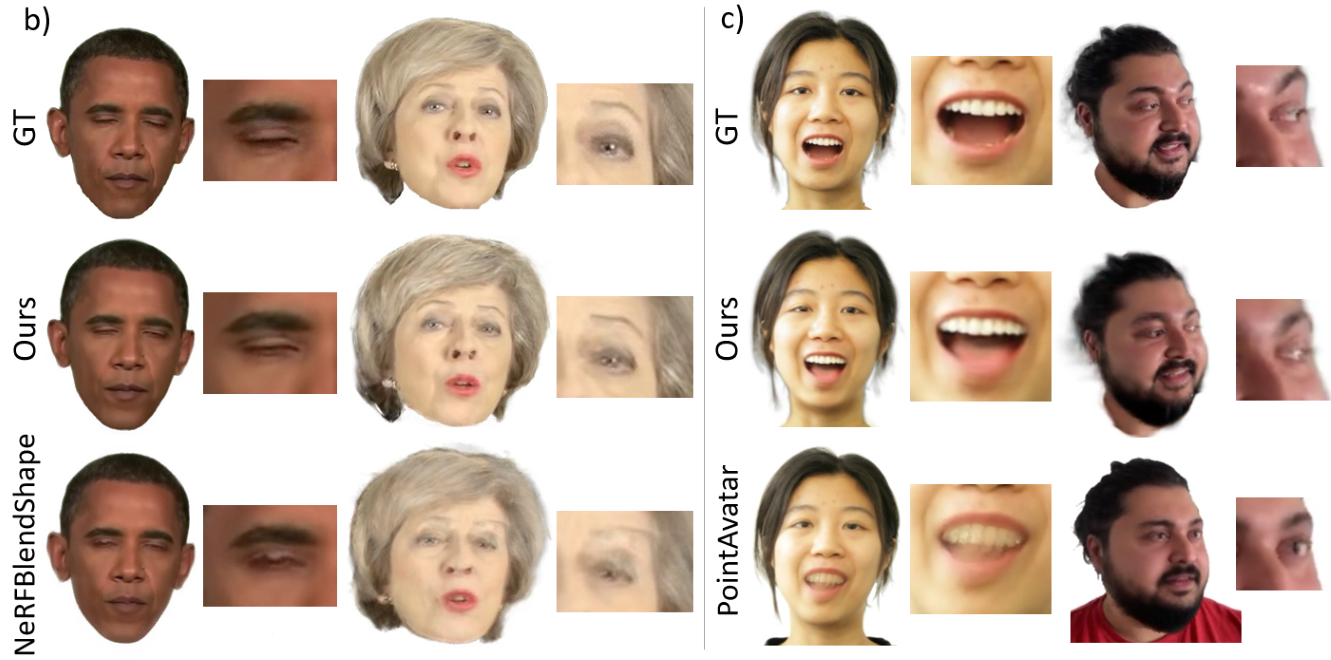} 
    \end{center}
    \caption{\textbf{Qualitative evaluation} comparing the proposed model against INSTA~\cite{INSTA:CVPR2023}, PointAvatar~\cite{Zheng2023pointavatar} and NeRFBlendShape~\cite{Gao2022nerfblendshape} baselines, namely on the \textbf{a)} INSTA data, \textbf{b)} NBS data and \textbf{c)} PointAvatar data. The close-ups on the right of each example highlight our method's ability to capture details like teeth, wrinkles and reflections.}
    \label{fig:result}
\end{figure*}

Table~\ref{tab:result1} reports the results of the metric comparison against baselines on the respective test sets. The three different blocks show namely results on the data released by INSTA, NBS and PointAvatar. Figure~\ref{fig:result} illustrates the qualitative comparison with the most recent baselines in all three datasets. %
We observe that the proposed method outperforms all baselines on the INSTA and NBS datasets in all metrics, with a PSNR gap of about $2$ dB. Referring to Figure~\ref{fig:result}, we observe that \methodNameSpace leads to higher fidelity to ground truth, less artifacts, and identity preservation for all subjects. Interestingly, as INSTA relies on mesh deformations, it exhibits artifacts such as noticeable triangles on the skin surface (Figure~\ref{fig:result}a).
Moreover, our model preserves better details, such as facial expressions, wrinkles, eyebrows, teeth and glass reflections, while other baselines~\cite{Zheng2023pointavatar, INSTA:CVPR2023,Gao2022nerfblendshape} often struggle in such aspects. The comparison against the PointAvatar baseline shows that we are superior on the INSTA data by about $2$ dB. On the 3 subjects of the PointAvatar dataset, our method surpasses the baseline in terms of LPIPS and SSIM, while having very similar PSNR. Overall, looking at both datasets, the performance of \methodNameSpace is superior to that of PointAvatar. Also qualitatively, we can see that PointAvatar results have distortions of some parts that undergo significant transformation, including inaccurate teeth deformation (Figure \ref{fig:result}c). We believe that, as an explicit method, PointAvatar generalizes well for under-observed expressions (resulting in comparable PSNR). However, as these cases struggle with deformation realism (teeth wrongly deform in the same way as the mouth), the structural metrics are worse. %
Finally, as the table shows, we improve the rendering time of all baselines for 512 resolution by at least a factor of 10. %
We refer the reader to the supplement for more qualitative results and videos.

\subsection{Ablations}
\label{sec:ablations}

In this section we ablate the \methodNameSpace components. First, we train a model that does not use learned features for blending, but rather a base of colors and opacities, and uses the expression weights to obtain the final color and opacity as a weighted average. We refer to this model as \textit{Ours w/o MLP}. In addition, since an intuitive alternative for dynamic Gaussians would be to deform the points (rather than adapting color and opacity) we introduce \textit{Ours w/ $\Delta(\mu,R)$}, which uses the learned feature basis (and a similar MLP) to rather shift the positions $\mu$ and rotations $R$. Additionally, we train a model that predicts color, opacity, as well as a $\mu$ and $R$ shift (\textit{Ours change all}). Further, we run a model without the perceptual loss, \ie \textit{Ours w/o $\mathcal{L}_{p}$}, to ablate its effectiveness. Additionally, to validate the contribution of using the expression parameters as a weight for blending Gaussian features, rather than a simple condition to the MLP, we ablate a variant named \textit{Ours w/o blending}. For fairness, we increase the capacity of this MLP until it plateaus. More details can be found in the supplement.

\begin{figure}[t]
    \begin{center}
        \includegraphics[width=0.88\linewidth]{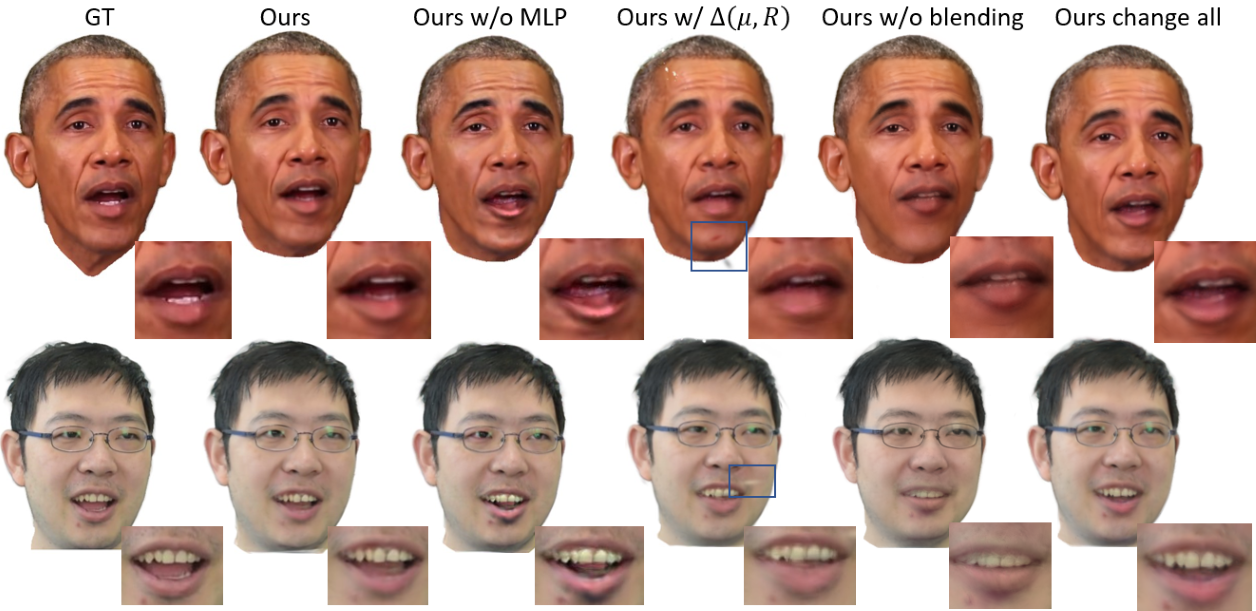} 
    \end{center}
    \caption{\textbf{Qualitative ablation on the INSTA dataset}}
    \label{fig:ablation}
\end{figure}

Table~\ref{tab:ablation} reports the quantitative evaluation of our model components. We observe that using the expression parameters as a simple condition (\textit{Ours w/o blending}) leads to noticeably inferior performance. In contrast to our model - which learns per-Gaussian dynamics via a feature basis - the per-subject MLP has to learn the face dynamics for all Gaussians at once, leading to poorer expression generalization, as can be also seen on Figure~\ref{fig:ablation}. Also, applying a transformation to the positions and rotations leads to worse results (\textit{Ours w/ $\Delta(\mu,R)$}). We hypothesize this is because, in the context of 3DGS - which is relying on several heuristics - adding another dimension (in the form of spatial 3D motion) further complicates the already difficult optimization, resulting in geometrically inconsistent transformation (\eg failure to preserve relative distances of points in the skin), especially for large motion. %
Figure~\ref{fig:ablation} confirms these results and reveals floater artifacts and less accurate expressions. We also observe that allowing all parameters to change (\textit{Ours change all}) increases the solution space and makes the heuristic-based 3DGS optimization more challenging, leading to blurrier results. Further, blending the explicit parameters directly (\textit{Ours w/o MLP}) leads to a worse performance than our neural variant. Despite a tighter PSNR gap, we notice a drastic visual effect on the highly dynamic areas, relevant to the blending, as illustrated in Figure \ref{fig:ablation}. Finally, we see that adding a perceptual loss term $\mathcal{L}_{p}$ leads to an improvement in most metrics. We refer the reader to the supplement for video comparisons, as well as ablation on additional aspects such as number of Gaussians and speed.

\begin{table}[t]
    \centering
    \caption{Ablation of \methodNameSpace components on the INSTA dataset.}
    \resizebox{0.57\columnwidth}{!}{%
    \begin{tabular}{l|c@{\hskip 0.1in}c@{\hskip 0.1in}c@{\hskip 0.1in}c@{\hskip 0.1in}c}
    \toprule
      Method   & L2 $\downarrow$  & PSNR $\uparrow$ & SSIM $\uparrow$ & LPIPS $\downarrow$ \\
      \midrule
     Ours \small{w/o blending}    & 0.0012 & 30.28  & 0.955  & 0.041 \\
        Ours \small{w/ $\Delta(\mu,R)$}    & 0.0014  & 29.83 & 0.953 & 0.045 \\
    Ours change all  & 0.0014  & 29.65  & 0.951  & 0.041 \\
    Ours \small{w/o MLP} & 0.0009 & 32.08 & 0.968 & \textbf{0.033} \\
     Ours \small{w/o $\mathcal{L}_{p}$}    & 0.0008 & 32.11 & 0.969 & 0.046 \\
     Ours    & \textbf{0.0008}  & \textbf{32.50} & \textbf{0.971} & \textbf{0.033} \\
     \bottomrule
    \end{tabular}
    }
    \label{tab:ablation}
\end{table}

\subsection{Novel View Synthesis}

\begin{figure}[t]
\begin{minipage}{.42\textwidth}
    \centering
    \includegraphics[width=0.98\linewidth]{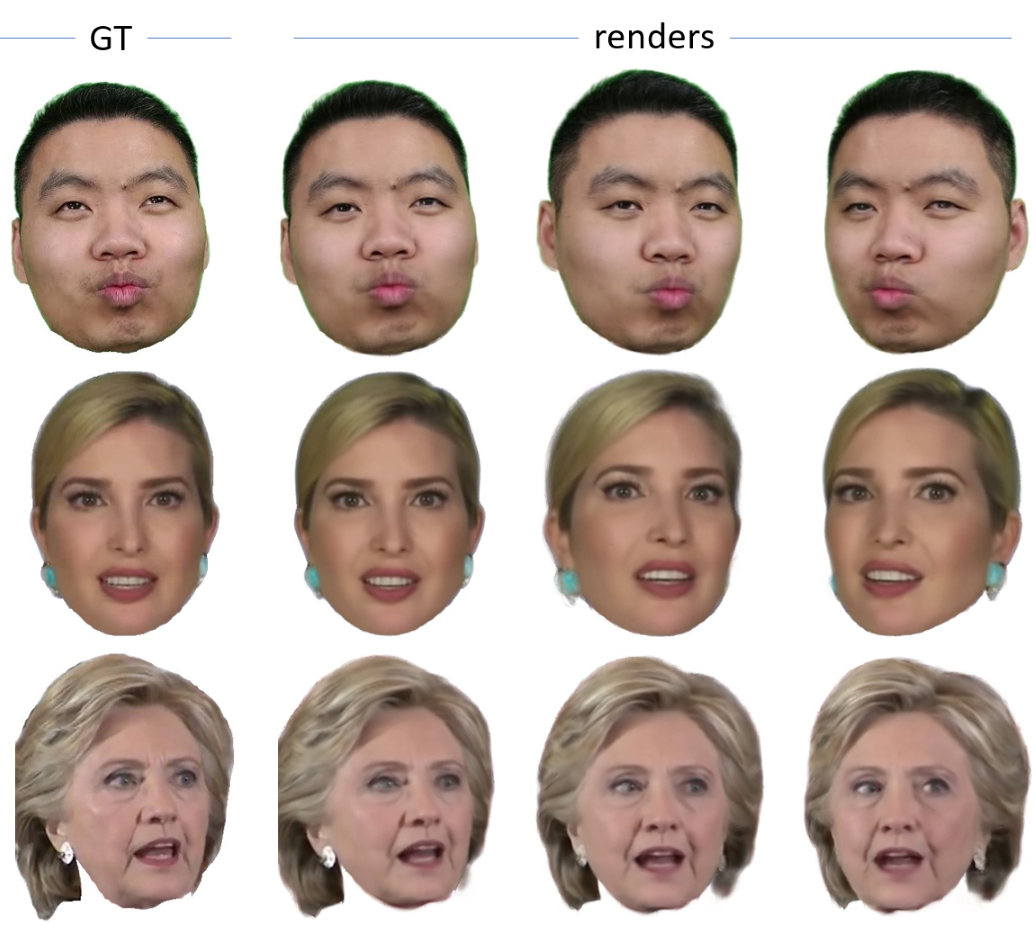}
    \caption{Rendering from various camera views for the same expressions}
    \label{fig:novelview}
\end{minipage}
\begin{minipage}{.54\textwidth}
    \centering
    \includegraphics[width=0.99\linewidth]{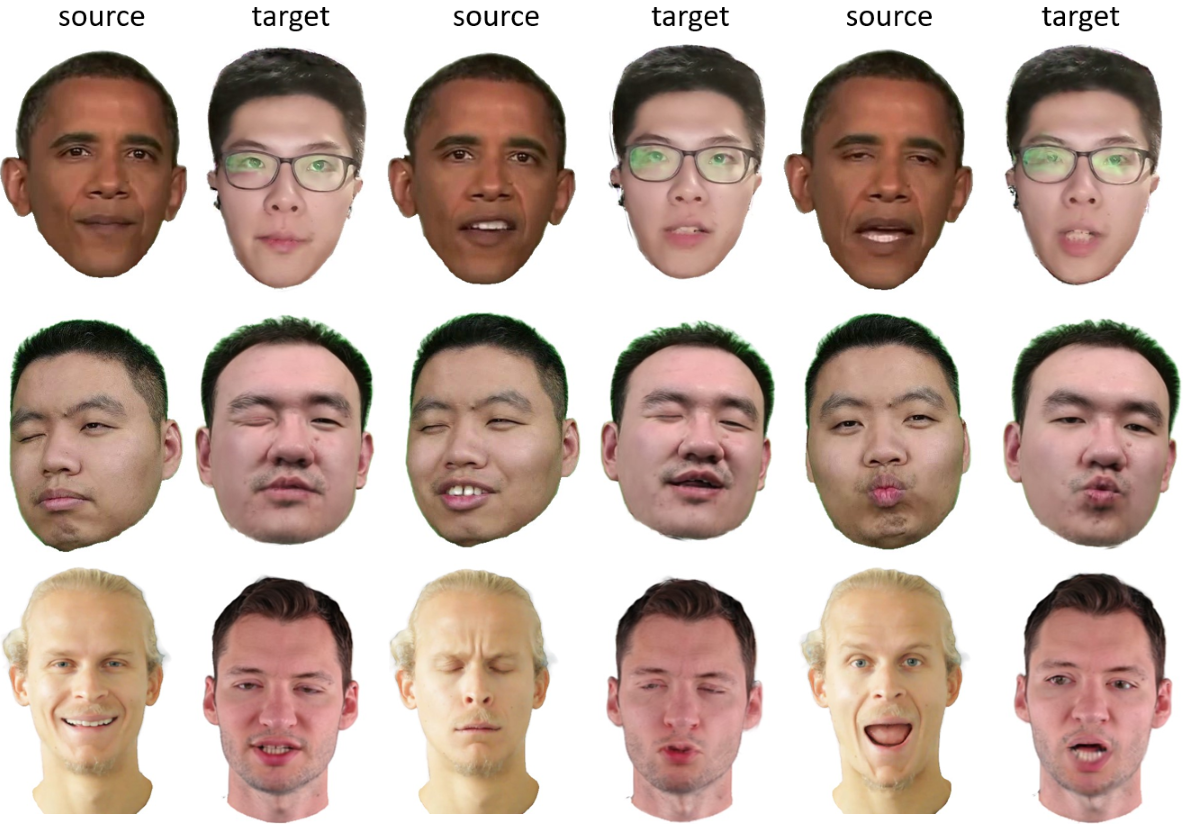}
    \caption{Cross-subject driving: rendering target subject with expression from the source subject}
    \label{fig:cross_driving}
\end{minipage}
\end{figure}

In Figure~\ref{fig:novelview} we render the avatars from multiple views, including the original test set camera (left) and two additional viewpoints (right). Thereby we render the same facial expression. We observe that our model can deliver expressions that are consistent across different views. Videos can be found in the supplement.  %

\subsection{Cross-Subject Expression Driving}

Figure~\ref{fig:cross_driving} reports our cross-subject driving results, \ie use the facial expression from another (ground truth) source subject to drive a target subject. Here we retain the original head pose of the target subject. As can be seen in the figure, our model is capable of transferring various expressions, \eg talking, wink, surprise, across different subjects at a reasonable quality. %

\begin{figure}[t]
    \centering
    \includegraphics[width=0.7\linewidth]{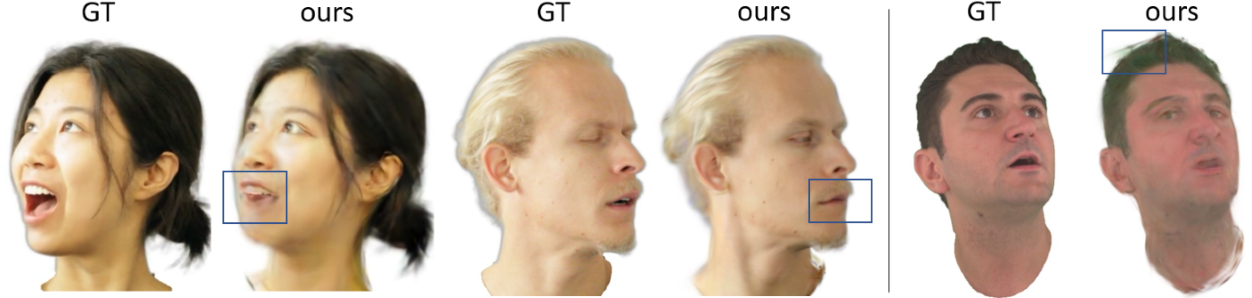}
    \caption{\textbf{Limitations.} \textit{Left:} subjects only observed in neutral expression in non-frontal view. \textit{Right:} Camera view that is far from the training observations.}
    \label{fig:limitations}
\end{figure}
\section{Limitations and Ethical Consideration}

\methodNameSpace is affected by failures of the head tracker, which can take the form of inaccurate facial expressions, or blurriness in case of pose inaccuracy in the training data (Figure~\ref{fig:limitations}). Further, as a data driven method, \methodNameSpace requires a reasonable coverage of expressions across different views. For instance, if a head changes pose only in neutral expression, and diverse expressions are observed only in frontal view, it would be difficult to capture a non-neutral expression from a side view (Figure~\ref{fig:limitations}, left). Finally, a downside of \methodNameSpace is the memory consumption originating from the feature bases ($B \times f_{dim}$ floats per Gaussian).  %

The use of personal data should be handled carefully, and follow local regulations.
We note that face reanimation can potentially be used to generate fake content, and convey misinformation. We do not condone such practises, and believe that the community should work accordingly towards mitigating the risks. 

\section{Conclusion}
\label{sec:conclusion}

We presented \methodName, a model for animatable head reconstruction and rendering from a monocular video that renders on real-time. Our extensive evaluation showed that the proposed model results in state-of-the-art performance, clearly surpassing the baselines, while rendering on real-time frame rates (about 250 fps for a $512^2$ resolution). We justified our design choices via a set of ablations, where we demonstrated that linearly blending implicit features leads to less artifacts than the alternative of blending explicit parameters directly. Moreover, we have shown that changing colors and opacity is more effective than the intuitive alternative of transforming the Gaussian mean positions. Future work can be dedicated to improving the memory efficiency of the \methodNameSpace feature bases. 

\bibliographystyle{eccv/splncs04}
\bibliography{main}
\clearpage

\section*{\\ \Large{Supplementary Material}} 

\renewcommand{\thesection}{\Alph{section}}
\setcounter{section}{0}

\setcounter{table}{0}
\renewcommand{\thetable}{S\arabic{table}}%
\setcounter{figure}{0}
\renewcommand{\thefigure}{S\arabic{figure}}%

This supplementary material provides further details about \methodName, as well as additional results. Section~\ref{sec:understanding} provides some insights on the learned 3D Gaussians, such as visualization of the learned feature basis as well as a gradual removal of the Gaussians to observe the occluded content. Section \ref{sec:supp_ablation} provides more implementation details on the methods used in the ablation of the main paper. Further, we provide some time analysis in Section \ref{sec:time}. In Section \ref{sec:supp_results} we show more qualitative results on the novel expression task comparing ours against baselines.
Finally, in this supplement we kindly refer the reader to a \textit{demo video} which contains method highlights and various result sequences as comparison with state-of-the-art methods, ablation and novel view synthesis.

\section{Understanding the Learned 3D Gaussians}
\label{sec:understanding}

\subsubsection{Basis visualization}

\begin{figure}[h!]
    \centering
    \includegraphics[width=0.5\linewidth]{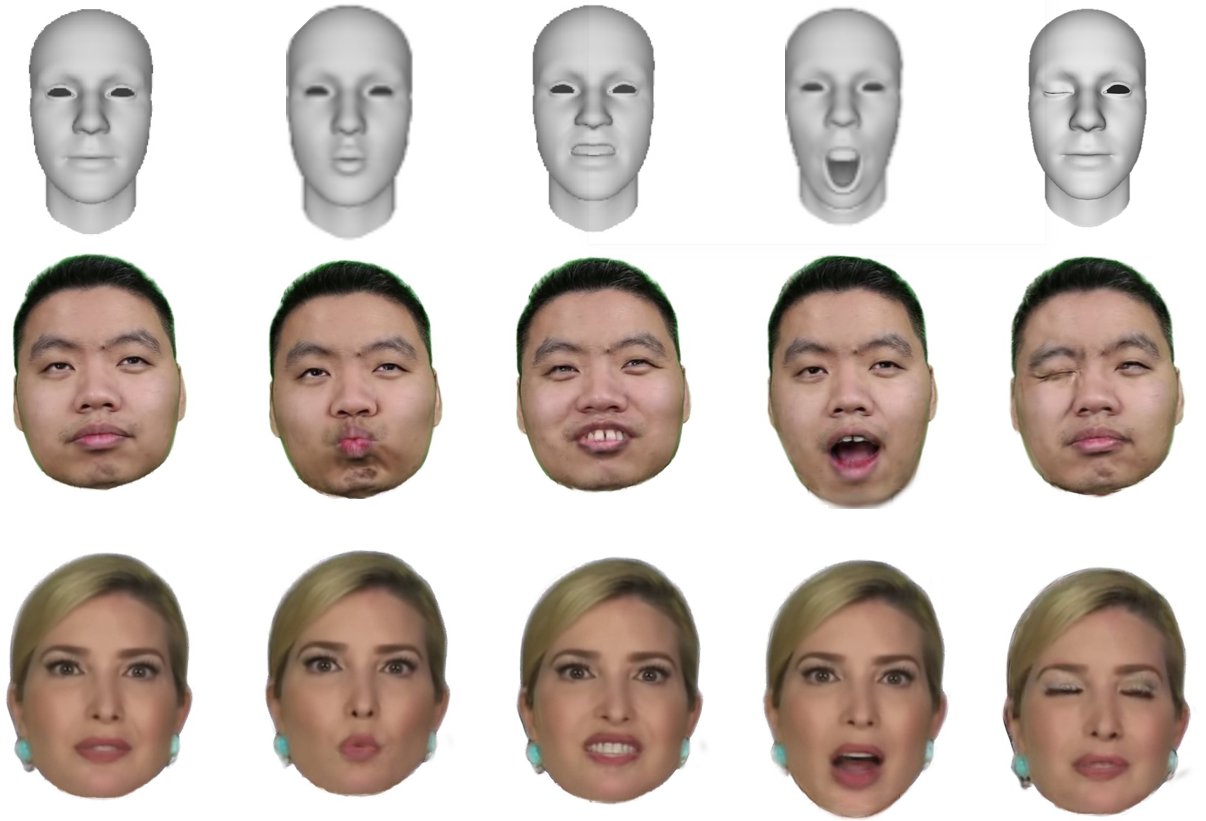}
    \includegraphics[width=0.47\linewidth]{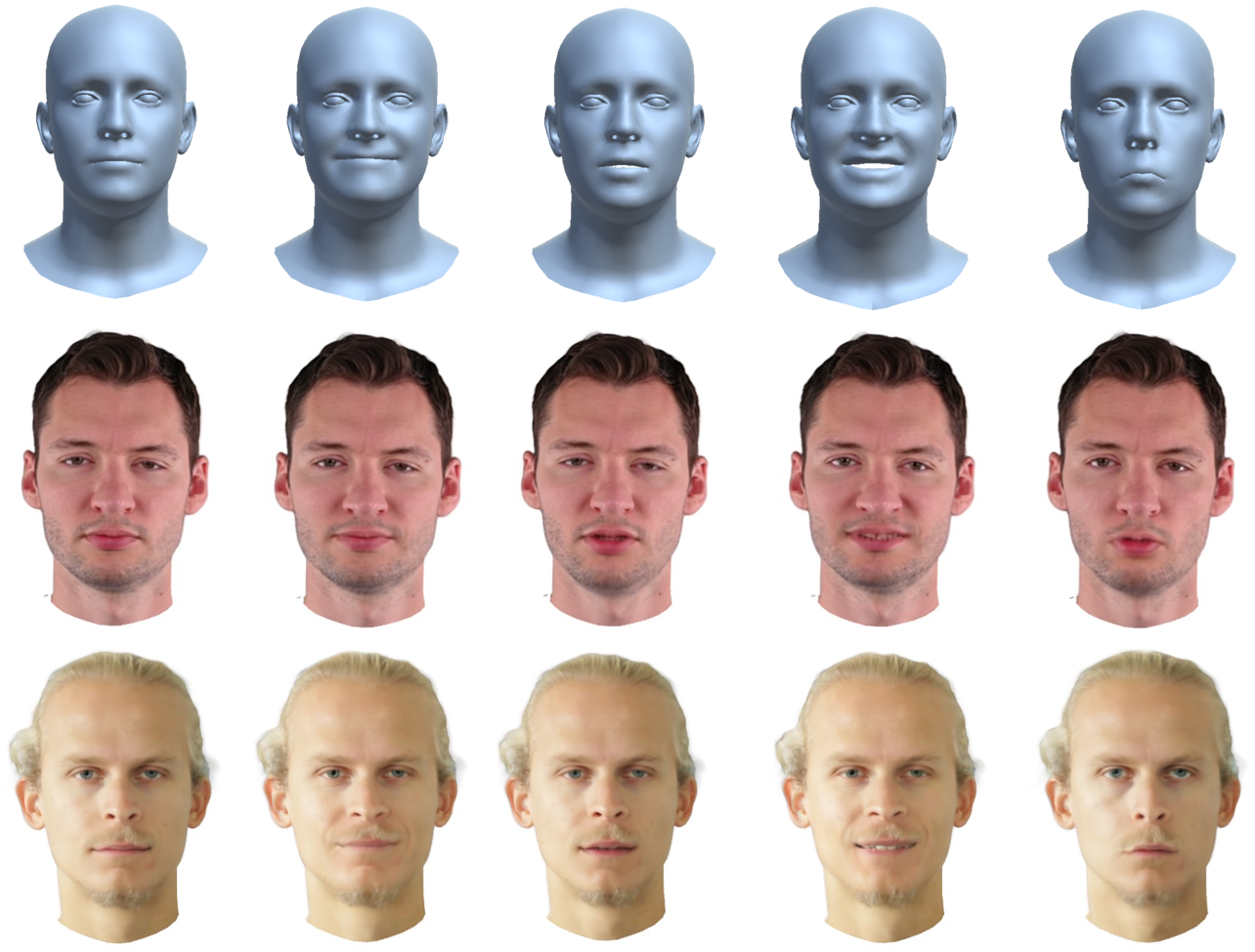}
    \caption{\textbf{Learned feature basis visualization.} \textit{Left:} Expression parameters from the Face Warehouse model \textit{Right:} Expression parameters from the FLAME model.}
    \label{fig:basis}
\end{figure}

 Since \methodNameSpace relies on a feature basis for blending, it would be beneficial to understand what this basis is learning. For this purpose, we utilize expression parameters as one-hot vectors to our model and show the generated image alongside the 3DMM mesh corresponding to that expression in Figure~\ref{fig:basis}. The left section of the figure shows two examples on the NBS data~\cite{Gao2022nerfblendshape}, using Face Warehouse~\cite{facewarehouse}, while the right part provides two FLAME~\cite{FLAME:SiggraphAsia2017} based examples from the INSTA data~\cite{INSTA:CVPR2023}. Note that for the FLAME-based tracking we noticed that the optimized neutral expressions were far from a vector of zeros, therefore we normalized the one-hot expression vectors first, by adding the expression weights of a neutral expression from the training set. Since the Face Warehouse basis is more semantic, \ie every expression element corresponds to a more local and interpretable action, such as winking, or opening mouth in surprise, we observe more drastic changes in this base compared to FLAME. The results show that \methodNameSpace learns a reasonable feature basis that aligns well with the 3DMM expressions. However, this is limited by the level in which an expression is observed in the training data, \eg for subject 2 from the top we observe that winking does not work quite well (both eyes are closing instead of one) due to this reason. 

\subsubsection{Gradual removal of Gaussians}
\begin{figure}[t]
    \centering
    \includegraphics[width=0.65\linewidth]{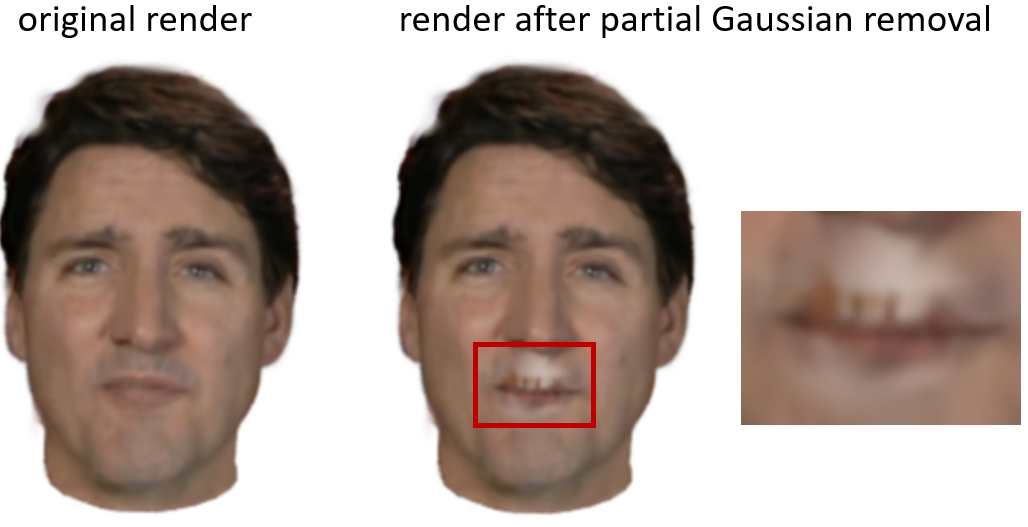}
    \caption{Removal of frontal 3D Gaussians reveals underlying structures such as teeth, invisible in this frame, but observed in other frames.}
    \label{fig:removal}
\end{figure}

The purpose of this experiment is to reveal what the intermediate Gaussians in a particular frame represent. This is of interest because the proposed method relies on over-representation, \ie multiple Gaussians will represent certain face areas (\eg lips) and they will occasionally become transparent to reveal other areas underneath (\eg teeth) as needed. Therefore we remove the optimized 3D Gaussians gradually, using the camera view as direction, from near to far. Figure \ref{fig:removal} shows the original rendering for a given frame as well as the rendering after we have removed the most frontal Gaussians in the mouth area. As expected, we start seeing teeth in the intermediate layers of visibility, as these structures have been observed from other frames (\eg when the person was talking or laughing). This reflects that, our model does not simply re-color the Gaussians in those areas to accommodate teeth instead of lips, but rather has a separate set of Gaussians to represent the teeth. Additionally, we observe background color (white) when dis-occluding the regions between the nose and the upper lip. This is due to the fact that there is no need to represent the structures underneath, as they are never observed by any frame.

\subsubsection{Effect of going outside of the training manifold}
\begin{figure}[t]
  \centering
\includegraphics[width=0.97\linewidth]{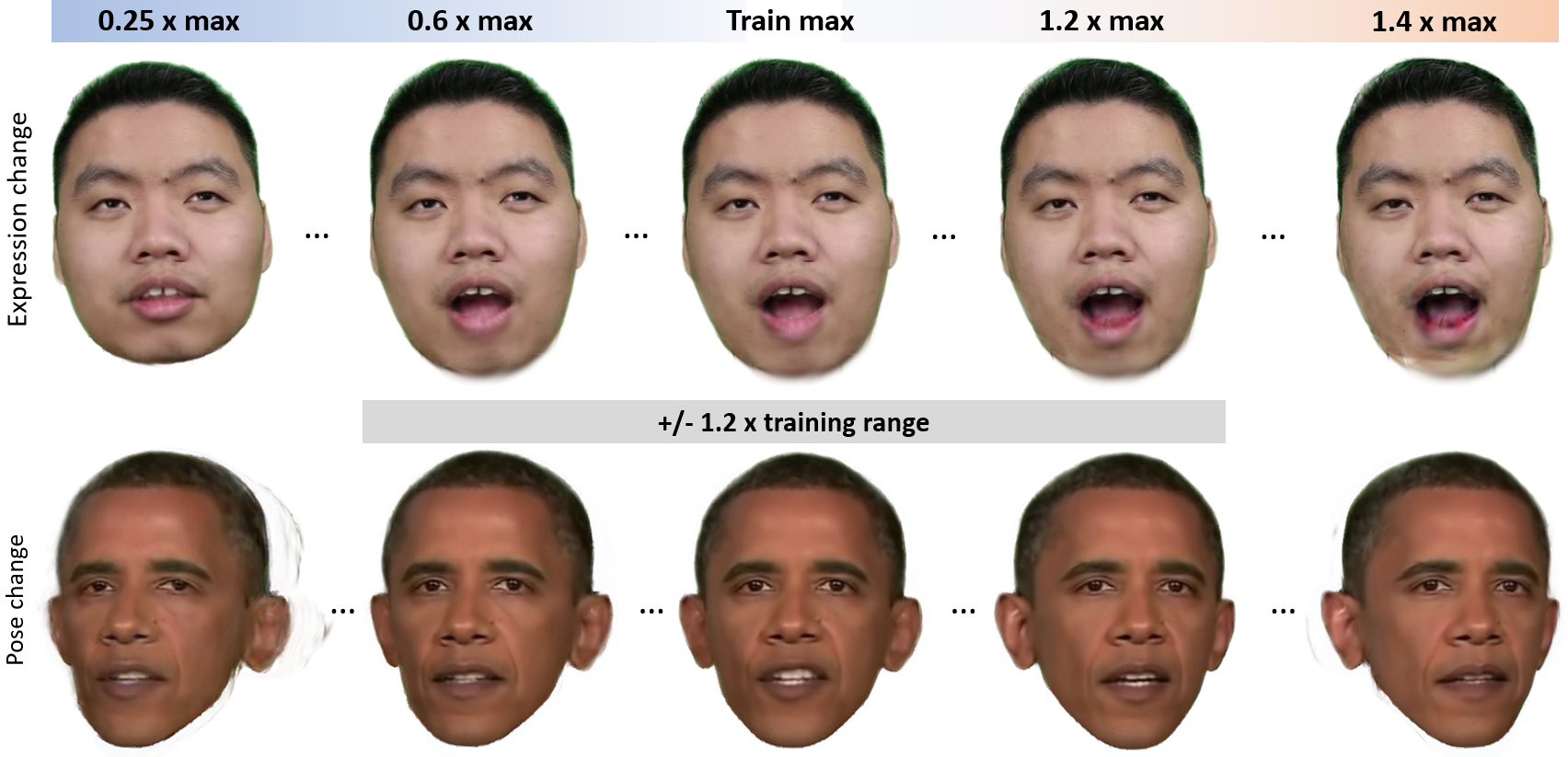}
   \caption{(Top) Navigation of 3DMM expression space for 1 parameter. (Bottom) Navigation of viewing angle.}
   \label{fig:navigation}
\end{figure}

In Figure~\ref{fig:navigation} (top) we navigate along one 3DMM parameter up to $1.4{\times}$ of its maximum value in the training set. We notice that colors start slowly to deteriorate after $1.2{\times}\rm{max}$. This observations is in line with the expectations, as we model deformation through changes in color and opacity. Similarly, in Figure~\ref{fig:navigation} (bottom) we see that if the viewing angle changes considerably from the range of observed training views, we start noticing some artifacts.

\section{Ablation Details}
\label{sec:supp_ablation}

This section provides more details on the ablation methods presented and evaluated in the main paper.

\subsubsection{Ours w/o MLP}
This model does not use learned features for the blending, but rather it blends a basis of colors and opacities directly. In each Gaussian we create these bases as $\bm{C} \in \mathbb{R}^{B \times 3(k+1)^2}$ and $\bm{\alpha} \in \mathbb{R}^{B \times 1}$ and use the expression weights to obtain the final color $\bm{c}_i$ and opacity $\alpha_i$ as a weighted average. Note that summation did not work here, as colors are explicit values and they would add up to high values.

\subsubsection{Ours w/ $\Delta(\mu,R)$}
This model uses the learned feature basis $\bm{F}$ of \methodNameSpace to rather shift the positions $\bm{\mu}$ and transform the rotations $\bm{R}$. Thereby we feed the blended feature $\bm{f}_i$ in to an MLP $\phi'(\cdot)$ that contains the same number of layers and hidden dimensions as $\phi(\cdot)$ from our proposed model. The difference is that, $\phi'(\cdot)$ outputs 3 values of position shift and 4 values of rotation (represented as quaternion) as:
\begin{equation}
    \Delta \bm{\mu}, \bm{r}_t = \phi'(\bm{f}_i, \psi(\bm{\mu})).
\end{equation}
These outputs are namely used to transform the static parameters of the Gaussian (after converting $\bm{r}_t$ to a rotation matrix $\bm{R}_t$) as %
\begin{equation}
    \bm{\mu}' = \bm{\mu} + \Delta \bm{\mu} 
\end{equation}
and
\begin{equation}
    \bm{R}' = \bm{R}_t \bm{R}.
\end{equation}
To facilitate convergence, we additionally add a regularization term to the estimated position shift, encouraging it to remain in a small range. The final loss then becomes
\begin{equation}
         \mathcal{L}_\text{total} = \lambda_{1} \mathcal{L}_{1}(I_\text{r}, I_\text{gt}) + \lambda_{s} \mathcal{L}_\text{SSIM} (I_\text{r}, I_\text{gt}) + \lambda_{p} \mathcal{L}_{p} (I_\text{r}, I_\text{gt}) + \lambda_{\mu} \mathcal{L}_1 (\Delta \bm{\mu}).
        \label{eq:loss_reg}
\end{equation}

\subsubsection{Ours change all} 
This model uses the learned feature basis $\bm{F}$ of \methodNameSpace to predict color, opacity as well as a shift of positions $\bm{\mu}$ and transformation of rotations $\bm{R}$. Thereby we feed the blended feature $\bm{f}_i$ to an MLP $\phi''(\cdot)$, which outputs sh colors, opacity, 3 values of position shift and 4 values of rotation (represented as quaternion). The transformations are applied in the same way as in \textit{Ours w/ $\Delta(\mu,R)$}.
Also here we apply the regularization term in the loss function, as in eq. (\ref{eq:loss_reg}).

\subsubsection{Ours w/o blending} This model aims to verify that using the expression parameters as a weight for blending Gaussian features works better than using it as a simple condition to the MLP. Therefore, here we do not have a basis of latent features $\bm{F}$. Instead, the expression vector $\bm{e}_i$ and the encoded position $\bm{\mu}$ are fed directly into $\phi(\cdot)$ to predict the color and opacity. We hypothesize that this baseline requires more capacity for the MLP, as, in contrast to our proposed method, it has to learn all dynamics of the face at once. Therefore, we do not restrict our experiments to a small MLP of two layers, but rather extend its capacity until it reaches a plateau. Thus, the MLP here results in five linear layers, each followed by a leaky ReLU.

\subsubsection{Ours w/o $\mathcal{L}_{p}$}
This model is the same as the proposed \methodNameSpace and simply has the perceptual loss disabled
\begin{equation}
        \mathcal{L}_\text{total} =  \lambda_{1} \mathcal{L}_{1}(I_\text{r}, I_\text{gt}) + \lambda_{s} \mathcal{L}_\text{SSIM} (I_\text{r}, I_\text{gt}).
\end{equation}

\section{Time analysis}
\label{sec:time}

\subsubsection{Rendering time vs image size}

\begin{figure}[h]
    \centering
    \includegraphics[width=0.75\linewidth]{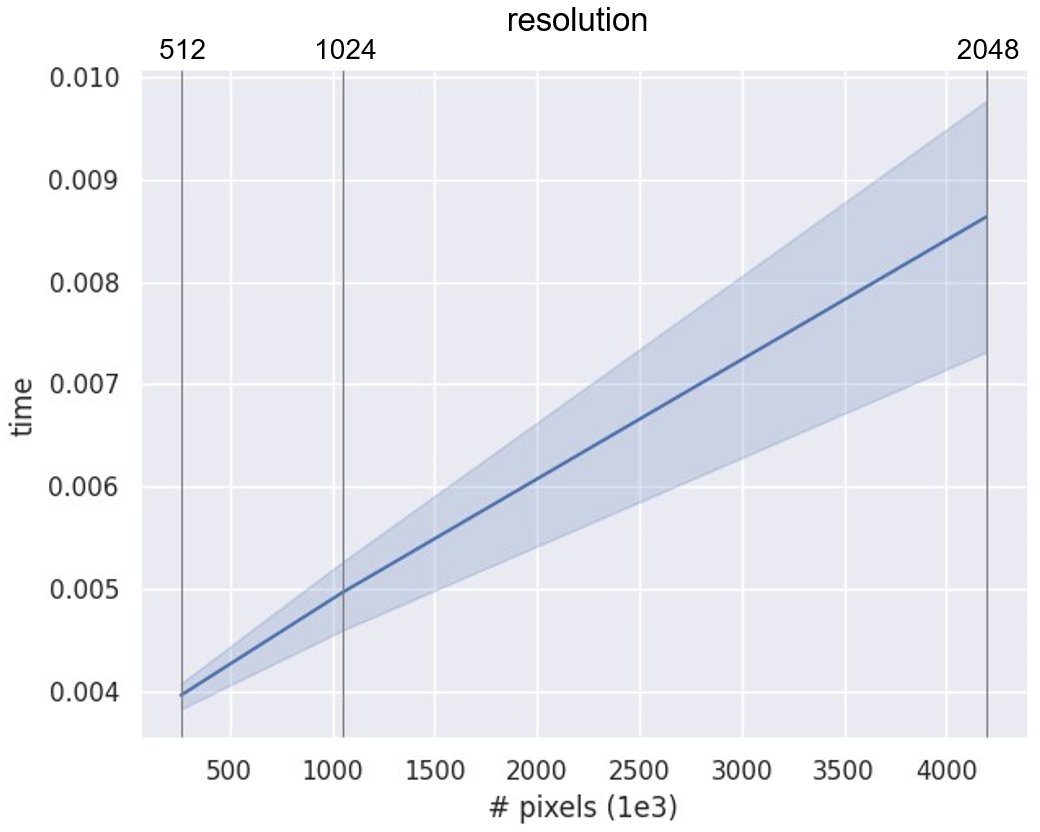}
    \caption{Rendering time versus resolution for \methodNameSpace on the INSTA dataset. We show the statistics of all subjects, with number of Gaussians ranging from 21k to 37k.}
    \label{fig:plot_res}
\end{figure}

In Figure \ref{fig:plot_res} we plot our models relationship between rendering time and image resolution. For each subject we render our models in 3 different resolutions, namely $512^2$, $1024^2$ and $2048^2$ and collect the run-time statistics. The number of Gaussians range between 21k and 37k, which is one of the main factors affecting the rendering time. The resulting mean rendering time for each resolution is namely 0.004, 0.005 and 0.0086, \ie the rendering time only doubles when we increase resolution by a factor of 4 in both dimensions (\ie 16$\times$ more pixels). 

\subsubsection{Ablation on number of 3D Gaussians and rendering time}
In addition to the image quality metrics evaluated in the main paper, we compare our different ablation models in terms of rendering time and number of Gaussians in Table \ref{tab:ablation_time}. We notice that the proposed \methodNameSpace (Ours) results in significantly less Gaussians compared to other models while being the fastest to render. Interestingly, the model that blends explicit parameters directly (Ours w/o MLP) is still slower than ours, despite not employing an MLP computation due to its large number of Gaussians. Another interesting observation is that, even though \methodNameSpace relies on over-representation, it still leads to significantly less Gaussians compared to the alternative model that transforms Gaussians (Ours \small{w/ $\Delta(\mu,R)$}). We believe this is due to the fact that the proposed model is more effective and easier to learn, and therefore it leads to the most efficient representation of space compared to other variants. 

\begin{table}[h!]
    \centering
    \caption{Ablation methods compared in terms of number of optimized 3D Gaussians and per-frame rendering time. Our method leads to the lowest number of Gaussians (\ie most efficient coverage of space) while having the best PSNR.}
    \resizebox{0.65\columnwidth}{!}{%
    \begin{tabular}{l|c@{\hskip 0.1in}c@{\hskip 0.1in}c}
    \toprule
      Method   & $\#$ Gaussians $\downarrow$  & Time (s) $\downarrow$ & PSNR $\uparrow$ \\
      \midrule
     Ours \small{w/o blending}    & 135k & 0.008 & 29.38  \\
     Ours change all & 97k & 0.012 & 29.65 \\
        Ours \small{w/ $\Delta(\mu,R)$}    & 234k & 0.019  & 29.83 \\
    Ours \small{w/o MLP} & 295k & 0.010 & 32.08 \\

     Ours    & \textbf{28k} & \textbf{0.004} & \textbf{32.50} \\

     \bottomrule
    \end{tabular}
    }
    \label{tab:ablation_time}
\end{table}

\section{Additional qualitative results}
\label{sec:supp_results}

\begin{figure}[h]
    \centering
    \includegraphics[width=0.6\linewidth]{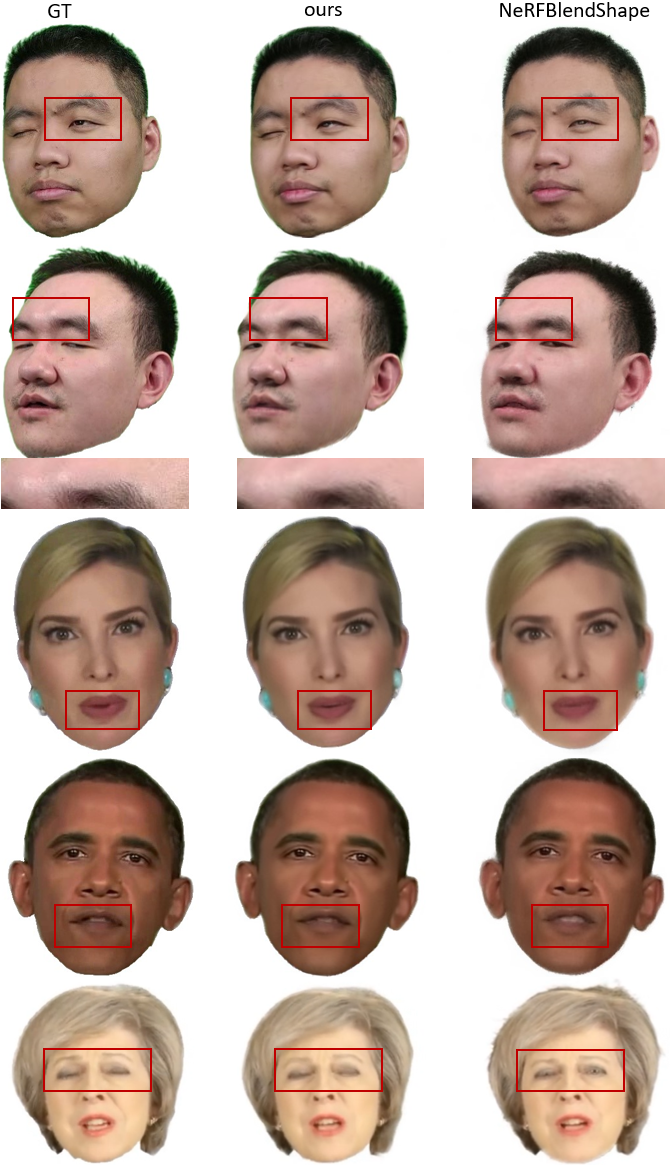}
    \caption{Additional qualitative results on the NBS data. \textit{Left:} Ground truth, \textit{Center:} \methodNameSpace (ours), \textit{Right:} NeRFBlendShape.}
    \label{fig:suppl_result_nbs}
\end{figure}

\begin{figure}[h!]
    \centering
    \includegraphics[width=0.99\linewidth]{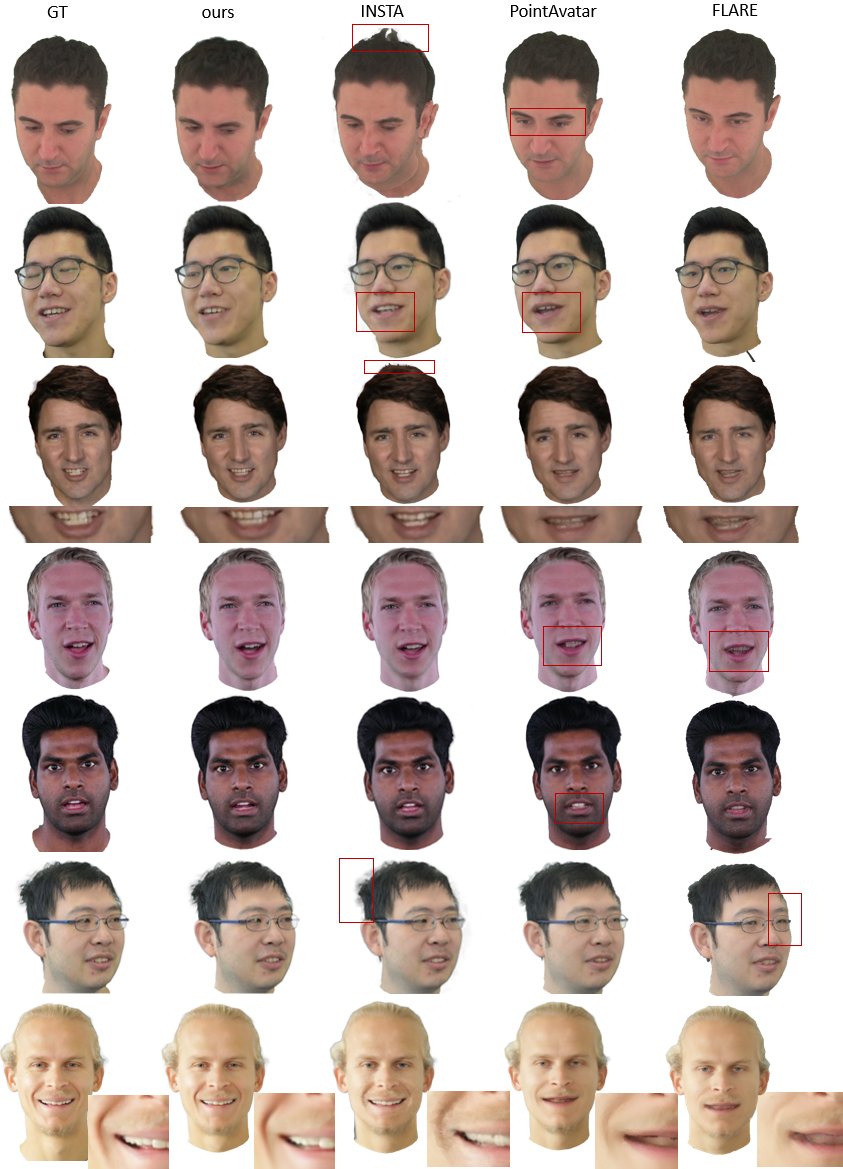}
    \caption{Additional qualitative results on the INSTA data. From left to right: Ground truth, \methodNameSpace (ours), INSTA,  PointAvatar and FLARE.}
    \label{fig:suppl_result}
\end{figure}

We provide additional qualitative results comparing \methodNameSpace against the most recent baselines \cite{INSTA:CVPR2023,Zheng2023pointavatar,Gao2022nerfblendshape,bharadwaj2023flare} in Figure~\ref{fig:suppl_result_nbs} and Figure~\ref{fig:suppl_result}. We observe that generally our model renders images with less artifacts, higher similarity to the ground truth expression, more noticeable reflecting glasses and skin specularities (Figure \ref{fig:suppl_result_nbs}, row 2).

\end{document}